\documentclass[draftcls,onecolumn]{IEEEtran}
\usepackage[T1]{fontenc}
\usepackage{graphicx}
\usepackage{booktabs}
\usepackage[misc]{ifsym}
\usepackage{bm}
\usepackage{amsmath}
\usepackage[utf8]{inputenc}
\usepackage{multirow}
\usepackage{algorithm}
\usepackage{algpseudocode}
\usepackage{orcidlink}
\usepackage{amssymb}
\usepackage{float}
\usepackage{adjustbox}
\newcommand{\x}{\mathbf{x}}
\newcommand{\y}{\mathbf{y}}
\newcommand{\A}{\mathbf{A}}
\newcommand{\I}{\mathbf{I}}

\newcommand{\z}{\mathbf{z}}
\newcommand{\w}{\mathbf{w}}
\newcommand{\D}{\mathbf{D}}
\newcommand{\bepsilon}{\bm{\epsilon}}

\newcommand{\md}{\mathrm{d}}
\hypersetup{hidelinks}

\begin{document}

\title{Tracing the Oracle: Improving Diffusion Timestep Scheduling for 3D CT Reconstruction}

\author{Yujia Wu and Zhaoqiang Liu$^{\ast}$\thanks{$^{\ast}$Corresponding author.} 
    
    \thanks{The authors are with the School of Computer Science and Engineering, University of Electronic Science and Technology of China.}}

\maketitle              

\begin{abstract}
Pretrained diffusion models demonstrate impressive potential in solving highly ill-posed 3D computed tomography (CT) inverse problems, while the inference process suffers from significant computational overhead. Furthermore, existing uniform timestep schedules fail to capture the non-uniform evolution of the reverse conditional diffusion stochastic differential equation, thereby introducing substantial truncation errors. To overcome this limitation, we propose Tracing the Oracle (TrO), a plug-and-play framework for improved timestep scheduling. Specifically, we treat densely sampled numerical integration trajectories on a few samples as the reference oracle. The optimized schedule is extracted by leveraging dynamic programming to globally minimize the cumulative error between the few-step approximation and the oracle. This mechanism precisely allocates the limited sampling steps to critical evolution stages that are highly susceptible to truncation errors. Our extensive experiments on the AAPM dataset across multiple 3D CT reconstruction tasks demonstrate that, when combined with the state-of-the-art 3D CT reconstruction method DDS, our optimized timesteps significantly improve reconstruction fidelity and computational efficiency compared to existing heuristic schedules, especially under a strict budget of no more than 10 sampling steps.
\end{abstract}

\section{Introduction}
The reconstruction of 3D computed tomography (CT) is mathematically formulated as a linear inverse problem. This problem aims to reconstruct an unknown clean signal $\x^*_0 \in \mathbb{R}^d$ from a degraded observation $\y \in \mathbb{R}^m$. This degradation process is mathematically represented by the linear forward operator $\A\in \mathbb{R}^{m\times d}$ and additive Gaussian noise $\bm{\xi}\in\mathbb{R}^m$ with $\bm{\xi}\sim \mathcal{N}(\mathbf{0}, \sigma_y^2\I_m)$, formalized as
\begin{equation}\label{eq:inverse_problem}
    \y = \A\x_0^* + \bm\xi.
\end{equation}

Deep neural networks trained on paired datasets have been widely employed to directly learn mappings from degraded observations to clean signals~\cite{dong2015image,tu2022maxim,zamir2021multi}. However, acquiring large-scale paired datasets is often prohibitively expensive, and such data-driven models exhibit severe generalization vulnerabilities across different measurement operators~\cite{delbracioinversion}. Deep generative models including variational autoencoders~\cite{kingma2013auto} and generative adversarial networks~\cite{goodfellow2014generative} have achieved remarkable success in solving inverse problems, by exploiting data-driven priors learned from unpaired training datasets~\cite{bora2017compressed,liu2022misspecified,chen2023unified,daras2021intermediate,chen2025solving}. Recently, diffusion models~\cite{ho2020denoising,song2021ddim,songscore} have emerged as powerful alternatives by modeling the data distribution through a stochastic differential equation (SDE). Leveraging pre-trained unconditional diffusion models as generative priors enables highly effective training-free inference for inverse problems~\cite{chung2023diffusion,zhu2023denoising,peng2024improving,zhang2024unleashing,chang2025provable,zhang2025improving,zheng2025integrating,li2026image,zheng2026image}. When addressing highly ill-posed 3D inverse problems, such as volumetric computed tomography (CT) reconstruction, directly training a 3D diffusion model imposes prohibitive computational costs~\cite{chung2023solving}. Consequently, existing diffusion posterior sampling approaches for 3D inverse problems primarily utilize 2D diffusion priors augmented with inter-slice regularization constraints~\cite{chung2023solving,chungdecomposed}. For instance, Decomposed Diffusion Sampling (DDS)~\cite{chungdecomposed} improves reconstruction efficiency by using CG-based data-consistency updates initialized from the Tweedie denoised estimate. Under the local assumption that the tangent space around the denoised estimate can be represented or approximated by a Krylov subspace, these updates remain within the corresponding local subspace~\cite{meurant2020krylov}. Nevertheless, existing approaches typically uniformly allocate the reverse sampling timesteps. Such heuristic timestep scheduling mechanisms fail to dynamically adapt to the complex interplay between the generative diffusion prior and the measurements of specific 3D inverse problems, thus yielding sub-optimal reconstruction quality under restricted sampling budgets. 

The sequential nature of the reverse diffusion process typically necessitates numerous neural function evaluations (NFEs), thereby imposing a substantial computational burden~\cite{lu2022dpmsolver,fu2026learnable,sun2026just}. To maximize sampling efficiency, various sampler optimization strategies have been proposed~\cite{frankel2025s4s,zhang2025lle}. In addition, various timestep scheduling optimization strategies have been proposed~\cite{xue2024optimizing,chen2024gits,zhou2024amed} to accelerate the sampling process. For instance, the work~\cite{xue2024optimizing} minimizes the explicit distance between numerical and analytic solutions, while GITS~\cite{chen2024gits} leverages the geometric regularity of probability flow trajectories. However, solving inverse problems relies on measurement-conditioned diffusion posterior sampling, where conditional measurements disrupt the unconditional generation manifold. Furthermore, achieving data consistency in inverse problems typically requires injecting stochastic noise during the reverse refinement process~\cite{zhu2023denoising,zhang2025improving}, which invalidates the deterministic trajectory assumptions relied upon by unconditional scheduling algorithms, rendering them ineffective for complex conditional inverse solvers.

\begin{figure}
    \centering
    \includegraphics[width=1.0\linewidth]{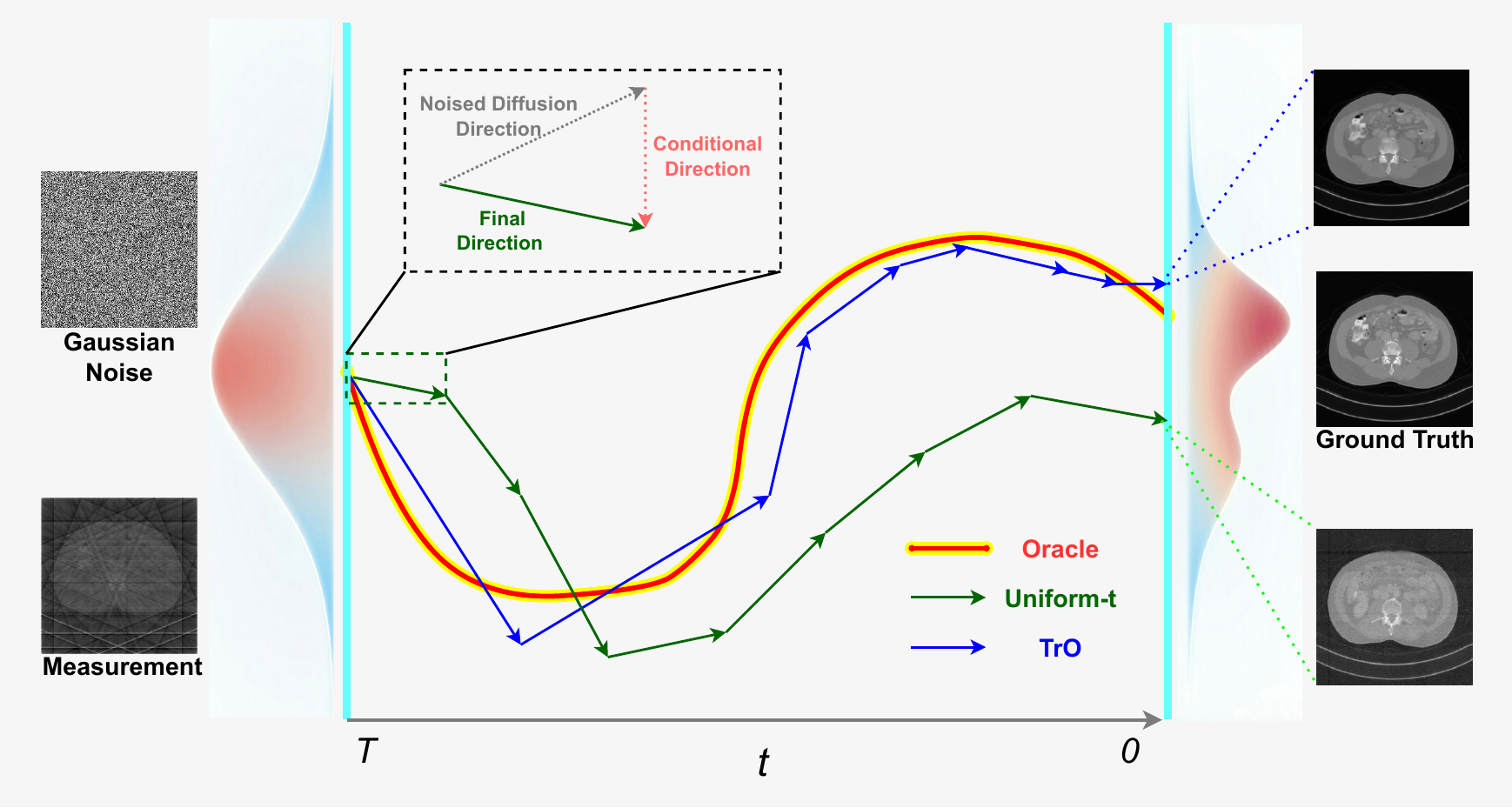}
    \caption{Geometric interpretation of the proposed Tracing the Oracle (TrO) framework. The heuristic uniform timestep schedule diverges from the high-fidelity oracle trajectory during the conditional generation process. Our method improves the timestep allocation by minimizing the cumulative approximation error to closely track the oracle path. This optimized non-uniform scheduling strategy concentrates limited computational resources on critical evolution stages and effectively approximates the ground truth.}
    \label{fig:headFig}
\end{figure}

To overcome these limitations, we propose Tracing the Oracle (TrO), which is a plug-and-play non-uniform timestep scheduling framework for 3D CT reconstruction (illustrated in Fig.~\ref{fig:headFig}). We initially construct a high-fidelity reference trajectory, referred to as the oracle, using a densely sampled stochastic process on a subset of data. To accurately quantify the approximation error across these stochastic transitions, we introduce a noise reuse technique that isolates the discretization error between arbitrary timesteps. Subsequently, we formulate the scheduling task as a discrete optimization problem and deploy dynamic programming to minimize the cumulative trajectory error between the few-step approximation and the reference oracle. Given that diffusion trajectories for identically distributed data exhibit strong structural similarities~\cite{chen2024gits}, the optimized timestep schedule generalizes robustly across the entire dataset without incurring additional inference overhead. This optimization framework enables highly efficient and accurate 3D CT reconstructions.

The primary contributions of this work are summarized as follows:
\begin{itemize}
    \item We propose a plug-and-play framework that formulates timestep scheduling for measurement-conditioned diffusion as a discrete shortest-path problem, aiming to minimize cumulative trajectory errors in 3D inverse problems.
    \item We introduce a noise reuse mechanism that isolates deterministic truncation errors from stochastic variance, enabling precise quantification of discretization errors within SDE-based sampling trajectories.
    \item Extensive evaluations on 3D CT benchmarks demonstrate that our optimized schedules consistently improve reconstruction fidelity over predefined schedules without additional inference overhead.
\end{itemize}

\label{sec:intro}

\section{Related Works}
\subsection{Diffusion Models for 3D Inverse Problems}
Medical image reconstruction requires solving mathematically ill-posed linear inverse problems, under severe measurement constraints driven by patient safety protocols~\cite{askari2025bgdm}. Diffusion priors have recently emerged as a prominent paradigm to regularize these underdetermined systems by enabling zero-shot inference over complex anatomical data manifolds~\cite{chung2023solving}. While these zero-shot conditioning methods successfully steer initial unconditional noise vectors toward measurement consistency, they inherently rely on reversing SDEs, which requires thousands of sequential integrations~\cite{song2023solving}. When scaled to high-dimensional volumetric domains such as 3D CT, these numerical integrations incur severe memory consumption and substantial inference latency, making existing frameworks computationally demanding~\cite{song2024diffusionblend,chung2023solving}.

To mitigate the memory limitations arising from native volumetric processing, recent literature explores patch-based and multi-view slice approximations. For instance, Song et al.~\cite{song2024diffusionblend} propose learning 3D patch image priors through position-aware diffusion score blending to ensure cross-slice consistency without relying on heuristic regularizers. Further compressing the representational space, He et al.~\cite{he2025blaze3dm} introduce a triplane neural field integrated with diffusion priors to enable fast zero-shot reconstruction while managing volumetric complexities. To enforce measurement fidelity, Askari et al.~\cite{askari2025bgdm} construct a bi-level guided diffusion model that approximates an inner-level conditional posterior mean before applying an outer-level proximal optimization.

To circumvent the substantial computational burden of computing manifold-constrained gradients, Chung et al.~\cite{chungdecomposed} exploit the observation that, if the tangent space at a Tweedie denoised sample is locally represented by a Krylov subspace, then CG updates initialized from the denoised estimate remain in that local tangent subspace. By executing classical conjugate gradient (CG) operations strictly within this clean tangent space, the framework significantly reduces the required NFEs while preserving high-fidelity volumetric spatial features. However, DDS typically follows a fixed uniform timestep schedule, which can lead to suboptimal reconstruction fidelity when restricted to a limited computational budget (i.e., the few-step regime). Our proposed method can improve the timestep schedule to further enhance the sampling efficiency and reconstruction performance of the DDS framework.

\subsection{Sampling Acceleration for Unconditional Diffusion}
To improve inference speed, researchers have focused on refining numerical solvers for diffusion-based differential equations. DDIM~\cite{song2021ddim} reformulates the generation pipeline from Markovian stochastic processes to deterministic numerical integration, allowing substantial reductions in temporal evaluations by skipping intermediate noise scales. DPM-Solver~\cite{lu2022dpmsolver} leverages the inherent semi-linear nature of probability flow ODEs by employing exact exponential integrators to compute linear drifts analytically while approximating nonlinear network predictions via high-order Taylor expansions. UniPC~\cite{zhao2023unipc} introduces a predictor-corrector framework that achieves high-order accuracy without requiring additional network evaluations per sampling step, significantly enhancing sample quality under limited sampling budgets.

In addition to solver design, recent studies focus on optimizing discretized timestep schedules and sampler coefficients to enhance low-budget generation. Sabour et al.~\cite{sabour2024ays} optimize non-uniform schedules by minimizing a Kullback-Leibler upper bound between exact and approximate paths. Xue et al.~\cite{xue2024optimizing} formulate a differentiable surrogate objective solved via trust-region methods to minimize the explicit distance between the numerical and analytical solutions. Zhou et al.~\cite{zhou2024amed} confirm that high-dimensional trajectories inherently reside within strict two-dimensional subspaces and invoke the mean value theorem to distill exact intermediate scaling factors that eliminate polynomial truncation errors.

The most relevant work to our approach is perhaps GITS~\cite{chen2024gits}, which focuses on the concept of trajectory geometry. GITS characterizes shape regularity within the implicit denoising trajectory and deploys dynamic programming to align the timestep schedule with the underlying structure of the probability flow ODE. While GITS demonstrates improved performance in unconditional generation, applying such deterministic trajectory geometry directly to conditional diffusion sampling corresponding to 3D inverse solvers poses significant challenges. We highlight two primary distinctions between our proposed framework and GITS designed to address these limitations. 1) GITS assumes a trajectory governed solely by the unconditional data manifold. In inverse problems, however, measurement-informed updates introduce external constraints that modify the trajectory geometry~\cite{zhang2025lle}. A schedule optimized for unconditional priors may misalign with these conditional dynamics, especially when the measurement operator induces high curvature in the manifold transitions. Furthermore, our empirical results in Sec.~\ref{sec:FurtherAnalysis} demonstrate that even distinct inverse problems exhibit varying preferences for timestep allocation, suggesting that the sampling dynamics are inherently coupled with the measurement process rather than being governed solely by the data prior. 2) GITS fundamentally relies on probability flow ODEs to compute deterministic state transitions. However, high-fidelity inverse problem solvers typically require diffusion SDEs to prevent the generated samples from drifting off the data manifold~\cite{song2024diffusionblend,zhu2023denoising,chung2023solving}. Directly substituting SDE transitions into an ODE-based error metric, or computing naive geometric distances between stochastic SDE states, introduces severe random variance that biases the true trajectory error estimation. Our framework resolves this issue by proposing a noise reuse mechanism that accurately isolates the deterministic truncation error within the stochastic sampling procedure.

\section{Preliminaries}

\subsection{Diffusion Models}
Diffusion models establish a generative framework by progressively corrupting a data distribution into a tractable prior distribution through a continuous-time stochastic process. Let $\x_0 \in \mathbb{R}^d$ denote a clean data sample drawn from the underlying distribution $p_{\text{data}}$. The forward process over a continuous time variable $t \in [0, T]$ is governed by an SDE formulated as
\begin{equation}\label{eq:forward_sde}
    \md \x_t = f(t) \x_t \md t + g(t) \md \w_t,
\end{equation}
where $\x_t \in \mathbb{R}^d$ represents the state vector at time $t$ and $\w_t \in \mathbb{R}^d$ denotes the standard Wiener process.  The function $f(t)$ acts as the drift coefficient while $g(t)$ functions as the diffusion coefficient. For each time $t$, the transition kernel from the initial data $\x_0$ to the latent state $\x_t$ follows a Gaussian distribution defined by
\begin{equation}\label{eq:marginal_dist}
    q(\x_t | \x_0) = \mathcal{N}(\x_t; \alpha_t \x_0, \sigma_t^2 \I_d),
\end{equation}
where $\alpha_t$ and $\sigma_t$ are differentiable functions representing the scale and noise schedule. Specifically, $\alpha_t$ is strictly decreasing while $\sigma_t$ is strictly increasing over $t$. To ensure that the SDE in Eq.~\eqref{eq:forward_sde} indeed induces the transition distribution $q(\x_t | \x_0)$, the drift and diffusion coefficients are related to these functions according to~\cite{lu2022dpmsolver}:
\begin{equation}
    \label{eq:define_f_n_g}
    f(t)=\frac{\md \log \alpha_t}{\md  t},\quad g^2(t)=\frac{\md  \sigma_t^2}{\md  t} - 2\sigma_t^2\frac{\md \log \alpha_t}{\md  t}.
\end{equation}
The boundary condition at the terminal time $T$ ensures that the marginal distribution approximates a standard normal distribution characterized by $\x_T \sim \mathcal{N}(\mathbf{0}, \I_d)$. The progression of the forward process is often measured by one half of the logarithm of signal-to-noise ratio (SNR), which is a strictly decreasing function of time denoted by $\lambda_t = \log ({\alpha_t}/{\sigma_t})$.

To invert this forward transformation and recover the original data distribution, a neural network model $\bepsilon_\theta(\x_t, t)$ parameterized by $\theta$ is trained to predict the injected noise $\bepsilon \sim \mathcal{N}(\mathbf{0}, \I_d)$. The network parameters are optimized by minimizing the following objective: 
\begin{equation}\label{eq:training_objective}
     \mathbb{E}_{t\sim \mathcal{U}(0,T), \x_0\sim p_{\text{data}}, \bepsilon\sim\mathcal{N}(\bm 0,\I_d)} \left[ \| \bepsilon_\theta(\alpha_t\x_0 + \sigma_t\bepsilon, t) - \bepsilon \|_2^2 \right].
\end{equation}
Song et al.~\cite{songscore} show that the reverse-time diffusion process evolving from $T$ down to $0$ can be formulated as an SDE:
\begin{equation}
    \label{eq:inverSDE}
    \md \x_t = \left(f(t)\x_t - g^2(t)\nabla_{\x_t}\log p_t(\x_t)\right)\md t + g(t)\md \overline{\w}_t,
\end{equation}
where $\overline{\w}_t$ is a standard Wiener process in the reverse-time direction, $p_t$ denotes the marginal probability density of $\x_t$ at time $t$, and the score function $\nabla_{\x_t}\log p_t(\x_t)$ can be approximated via the noise prediction network as $\nabla_{\x_t}\log p_t(\x_t) \approx -\bepsilon_\theta(\x_t, t)/\sigma_t$. Furthermore, there exists a deterministic probability flow ODE that shares the same marginal densities as the SDE~\cite{songscore}:
\begin{equation}
    \label{eq:inverODE}
    \md \x_t = \left(f(t)\x_t - \frac{1}{2}g^2(t)\nabla_{\x_t}\log p_t(\x_t)\right)\md t.
\end{equation}
In practice, Eq.~\eqref{eq:inverODE} is integrated numerically over a user-defined fixed timestep schedule $\{t_i\}_{i=0}^N$, where $t_i>t_{i+1}$ for $i=0,1,...,N-1$ with $t_0 = T$ and $t_N = 0$. At an arbitrary timestep $t_i$, Tweedie's formula~\cite{efron2011tweedie} provides the expected posterior mean of the clean data, which is approximately derived as:
\begin{equation}\label{eq:tweedie_est}
    {\x}_{0|t_i} = \cfrac{\x_{t_i} + \sigma_{t_i}^2\nabla_\x\log p_{t_i}\left(\x_{t_i}\right)}{\alpha_{t_i}}\approx\cfrac{\x_{t_i} - \sigma_{t_i} \bepsilon_\theta(\x_{t_i}, t_i)}{\alpha_{t_i}}.
\end{equation}
The state transition from the current step $t_i$ to the subsequent step $t_{i+1}$ follows a deterministic numerical integration rule given by DDIM~\cite{song2021ddim} as:
\begin{equation}\label{eq:ddim_step}
    \x_{t_{i+1}} = \alpha_{t_{i+1}} {\x}_{0|t_i} + \sigma_{t_{i+1}} \bepsilon_\theta(\x_{t_i}, t_i).
\end{equation}
\subsection{Decomposed Diffusion Sampler (DDS)}
Training-free approaches solve the linear inverse problem in Eq.~\eqref{eq:inverse_problem} by drawing samples from the conditional posterior distribution. However, achieving high reconstruction fidelity with such training-free solvers~\cite{song2023solving,chung2023solving,chung2022improving} necessitates multiple NFEs. DDS~\cite{chungdecomposed} mitigates these computational burdens by performing data-consistency updates on the Tweedie denoised representation. The key idea is that, under a local Krylov-subspace approximation of the tangent space around ${\x}_{0|t_i}$, CG updates initialized from ${\x}_{0|t_i}$ provide an efficient local residual minimization strategy without explicitly computing manifold-constrained gradients. At timestep $t_i$, to address highly ill-posed 3D inverse problems such as volumetric CT reconstruction, the algorithm seeks a refined estimate $\hat\x$ by solving a regularized optimization problem:
\begin{equation}\label{eq:cg_optimization}
    \min_{\x} \frac{1}{2} \|\y - \A\x\|_2^2 + \zeta \|\D_z \x\|_1,
\end{equation}
where $\D_z$ denotes the finite difference operator along the z-axis for Total Variation regularization~\cite{sidky2008image}, and $\zeta>0$ controls the regularization strength. 

To efficiently solve this while constraining the update direction strictly along the appropriate manifold tangent space, the algorithm utilizes the Alternating Direction Method of Multipliers (ADMM)~\cite{boyd2010distributed}. By introducing an auxiliary variable $\z$ and a scaled dual variable $\w$, the optimization is decomposed into tractable subproblems. Specifically, the data consistency update for $\x$ requires solving a linear system governed by $\A_\text{CG} = \A^\top\A + \rho\D_z^\top\D_z$ and $\mathbf{b}_\text{CG} = \A^\top \y + \rho\D_z^\top(\z - \w)$, where $\rho>0$ is the ADMM penalty parameter enforcing the constraint, while the TV regularization strength is controlled by $\zeta$. This system is efficiently resolved by applying $M$ steps of the CG method~\cite{chungdecomposed} initialized at the unconditional Tweedie estimate ${\x}_{0|t_i}$ in Eq.~\eqref{eq:tweedie_est}:
\begin{equation}\label{eq:cg_update}
    \hat{\x}_{0|t_i} = \text{CG}(\A_\text{CG}, \mathbf{b}_\text{CG}, {\x}_{0|t_i}, M).
\end{equation}
Subsequently, the new auxiliary variable $\z'$ is updated using a soft-thresholding operator $\mathbf{S}_{\zeta/\rho}(\cdot)$ as follows:
\begin{equation}
    \z'=\mathcal{S}_{\zeta/\rho}(\D_z\hat{\x}_{0|t_i}+\w),
\end{equation}
where $\mathcal{S}_{\omega}(u) = \text{sign}(u) \cdot \max(|u| - \omega, 0)$.
In addition, the new scaled dual variable $\w'$ is updated as follows:
\begin{equation}
    \w'=\D_z\hat{\x}_{0|t_i}+\w-\z'.
\end{equation}
Implementation details of the ADMM and CG processes are provided in the supplementary material. Following this refinement, the unconditioned estimate in the discrete integration rule is substituted by the data-consistent estimate $\hat{\x}_{0|t_i}$. The algorithm transitions from the current timestep $t_i$ to the subsequent step $t_{i+1}$ utilizing a stochastic generalized timestep schedule as DDIM~\cite{song2021ddim}, rather than the purely deterministic integration rule. The transition incorporates the measurement information defined by
\begin{equation}\label{eq:dds_transition}
    \x_{t_{i+1}} = \alpha_{t_{i+1}} \hat{\x}_{0|t_i} + \sqrt{\sigma_{t_{i+1}}^2 - \eta^2 {\beta}_{t_{i+1}}^2} \bepsilon_\theta(\x_{t_i}, t_i) + \eta {\beta}_{t_{i+1}} \bepsilon_i,
\end{equation}
where $\bepsilon_i \sim \mathcal{N}(\mathbf{0}, \I_d)$ represents standard Gaussian noise. Here, $\eta \in [0, 1]$ regulates the degree of stochasticity and $\beta_{t}$ represents the standard deviation of the noise at time $t$ as defined in DDIM~\cite{song2021ddim}. The deterministic trajectory is susceptible to accumulated approximation errors from the iterative data consistency projections. Introducing controlled stochastic noise mitigates these deviations and functions as a corrective mechanism to prevent samples from drifting off the clean data manifold.

\section{Methods}
To further enhance the sampling efficiency of diffusion-based 3D inverse solvers and enable high-fidelity reconstruction within the few-step regime, we propose Tracing the Oracle (TrO). Our framework formulates improved timestep scheduling as a discrete shortest-path optimization problem, identifying an optimized subset of evaluation points that minimizes the global truncation error for 3D inverse problems.

\subsection{Oracle Trajectory}
Continuous reverse diffusion processes trace trajectories along the data manifold from noise to real data. In the few-step regime, approximating this continuous evolution with sparse discrete transitions introduces truncation errors, causing the generated samples to deviate from the true data manifold and affecting reconstruction fidelity. 

Conversely, numerical integration with dense evaluations more closely tracks the continuous conditional distribution. Therefore, we utilize DDS~\cite{chungdecomposed} with a large number of evaluations $N$ to generate a reference trajectory for a given measurement $\y$. We define this densely sampled path as the oracle trajectory, denoted by $\mathcal{T}_{\text{oracle}} = \{\x_{t_i}\}_{i=0}^N$. Our objective is to extract a target subset of $L$ timesteps, $\{\tau_j\}_{j=0}^{L}$ ($L \ll N$), such that the trajectory generated by this subset minimizes the cumulative approximation error relative to the oracle under our specified metric.

\subsection{Discretization Error}
To extract the optimized subset, it is necessary to quantify the transition error incurred by skipping intermediate timesteps. Suppose we transition from a source timestep $t_i$ to a target timestep $t_j$ ($t_i > t_j$), following the DDIM sampling rule in Eq.~\eqref{eq:dds_transition}, the one-step approximation $\overline{\x}_{t_j|t_i}$ is derived as:
\begin{equation}\label{eq:j2iInverSample}
    \overline\x_{t_{j}|t_{i}} = \alpha_{t_j}\hat{\x}_{0|t_i} + \sqrt{\sigma^2_{t_j}-\eta^2{\beta}^2_{t_j}}\bepsilon_\theta(\x_{t_i}, t_i) + \eta{\beta}_{t_j}\bepsilon,
\end{equation}
where $\hat{\x}_{0|t_i}$ is the data-consistent estimate obtained through the CG optimization at time $t_i$ in Eq.~\eqref{eq:cg_update}, and $\bepsilon \sim \mathcal{N}(\mathbf{0}, \I_d)$ is the injected standard Gaussian noise. The discretization error associated with this transition can be evaluated by measuring the geometric distance $\|\overline{\x}_{t_j|t_{i}}-\x_{t_j}\|_2$ between the approximation $\overline{\x}_{t_j|t_i}$ and the corresponding intermediate state $\x_{t_j}$ from the oracle trajectory following GITS~\cite{chen2024gits}.

\subsection{Noise Reuse}
However, the geometric distance between $\overline{\x}_{t_j|t_{i}}$ and $\x_{t_j}$ presents challenges in stochastic frameworks. Since both the dense oracle trajectory and the one-step approximation independently sample the stochastic noise term $\bepsilon$, the resulting difference $\overline{\x}_{t_j|t_{i}} - \x_{t_j}$ is non-deterministic. Even if the deterministic components are fixed, independent target-step noise injections contribute
\[
\mathrm{Cov}(\eta\beta_{t_j}\bepsilon - \eta\beta_{t_j}\bepsilon') 
= 2\eta^2\beta_{t_j}^2\mathbf{I}_d,
\]
where $\bepsilon$ and $\bepsilon'$ are independently sampled standard Gaussian noises.
This covariance indicates that the geometric distance fluctuates, which obscures the deterministic truncation error. While setting the stochasticity parameter $\eta$ to zero avoids this variance, executing purely deterministic sampling in conditional inverse solvers can trap the generation process in local optima and degrade reconstruction performance~\cite{zhang2025improving,chungdecomposed,zhu2023denoising}. 

To address this, we employ a noise reuse mechanism. During the generation of the oracle trajectory $\mathcal{T}_{\text{oracle}}$, we explicitly record the stochastic noise vectors $\bepsilon_i$ injected at each step $t_i$. Specifically, in the error evaluation phase, to compute the approximation state $\overline{\x}_{t_j|t_{i}}$ targeting the oracle state $\x_{t_j}$, we reuse the exact noise vector $\bepsilon_{j-1}$ that was originally utilized during the oracle transition from $t_{j-1}$ to $t_j$. This precise alignment neutralizes the random variance and ensures that the approximated state remains consistent with the marginal distribution of the actual SDE sampling trajectory, thereby isolating the deterministic truncation error within the stochastic sampling procedure.

Consequently, we define the transition cost $C_{i,j}$ from $t_i$ to $t_j$ as:
\begin{equation}\label{eq:cost_matrix}
    C_{i,j} = \|\overline{\x}_{t_j|t_{i}} - \x_{t_j}\|_2 + \kappa \left( (t_i - t_j) - \frac{T}{L} \right)^2.
\end{equation}
The first term represents the truncation error isolated via noise reuse. The second term is a regularization component scaled by a hyperparameter $\kappa \geq 0$. This regularization penalizes large deviations from a uniform temporal stride ($T / L$) to prevent the neural network from operating outside its trained conditional manifold. The computational procedure for constructing the cost matrix is detailed in Algorithm~\ref{alg:TrO}.

\subsection{Schedule Optimization}
Given the cost matrix $\mathbf{C} \in \mathbb{R}^{(N+1) \times (N+1)}$ and $\{t_i\}_{i=0}^N$, the timestep scheduling task is formulated as a discrete shortest path problem to find the optimized timestep schedule $\{\tau_{j}\}_{j=0}^{L}$. Our objective is defined as:
\begin{equation}\label{eq:optimModel}
    \begin{aligned}
        \min_{\{\tau_{j}\}_{j=0}^{L}} \quad & \sum_{j=0}^{L-1} C_{k_j, k_{j+1}}, \\
        \text{s.t.}\quad & \tau_j=t_{k_j},\ k_j \in \{0, 1, \dots, N\},\quad j=0,1,...,L\\
        & \tau_j>\tau_{j+1}, \quad j=0, 1, \dots, L-1,\\
        & \tau_0=t_0,\tau_{L}=t_N.
    \end{aligned}
\end{equation}

To compute a schedule that minimizes this objective over the directed acyclic graph defined by $\mathbf{C}$, we implement the dynamic programming algorithm (DP)~\cite{bellman1966dynamic} similar to GITS~\cite{chen2024gits}, presented in Algorithm~\ref{alg:dp}. Specifically, the DP state records the minimum accumulated cost of reaching a timestep with a fixed number of transitions by enumerating all possible predecessor nodes. Therefore, given the pre-computed cost matrix $\mathbf{C}$, the returned schedule is the global optimum of Eq.~\eqref{eq:optimModel} rather than a greedy or locally optimized solution. Without dynamic programming, exact search over all feasible schedules leads to a combinatorial time complexity of \(\mathcal O\!\left(L\binom{N-1}{L-1}\right)\). Given the precomputed cost matrix \(\mathbf C\), our DP solver instead obtains the global optimum in \(\mathcal O(LN^2)\) time and \(\mathcal O(LN)\) memory.

Data samples originating from the same underlying distribution typically exhibit consistent structural properties within their reverse diffusion trajectories~\cite{chen2024gits,frankel2025s4s}. Thus, the timestep schedule extracted from a subset of calibration data generalizes across the test dataset. In our pipeline, the reference trajectory generation and the DP search are conducted in an offline preparatory phase. During the inference stage for 3D volume reconstruction, the pre-computed schedule $\{\tau_j\}_{j=0}^{L}$ is applied as a target timestep schedule, ensuring that this scheduling approach introduces no additional computational overhead to the conditional generation process.

\begin{algorithm}[!t]
    \caption{Tracing the Oracle (TrO)}
    \label{alg:TrO}
    \begin{algorithmic}[1]
        \Require Source timesteps $\{t_i\}_{i=0}^{N}$, $\{\alpha_{t_i}\}_{i=0}^{N}$, $\{\sigma_{t_i}\}_{i=0}^{N}$ and $\{\beta_{t_i}\}_{i=0}^{N}$, target number of steps $L$, regularization parameter $\kappa$, $\A$, $\y$, $\D_z$, $\rho$, $\mathbf{S}_{\zeta/\rho}$, $M$, $\eta$
        \State Sample $\x_{t_{0}} \sim \mathcal{N}(\mathbf{0}, \mathbf{I}_d)$
        \State $ \z_0 \gets \mathbf{0}_{d}, \w_0 \gets \mathbf{0}_d$
        \For {$i \gets 0$ to $N-1$}
            \State $\hat\bepsilon_i \gets \bepsilon_\theta(\x_{t_i}, t_i)$
            \State $\x_{0|t_{i}} \gets (\x_{t_i} - \sigma_{t_i}\hat\bepsilon_i)/\alpha_{t_i}$ \Comment{Tweedie denoising}
            \State $\A_\text{CG} \gets \A^\top\A + \rho\D_z^\top\D_z$
            \State $\mathbf{b}_\text{CG} \gets \A^\top \y + \rho\D_z^\top(\z_{i}-\w_{i})$
            \State $\hat{\x}_{0|t_{i}} \gets \text{CG}(\A_\text{CG}, \mathbf{b}_\text{CG}, \x_{0|t_{i}}, M)$ \Comment{Data consistency}
            \State $\z_{i+1} \gets \mathbf{S}_{\zeta/\rho}(\D_z\hat{\x}_{0|t_{i}} + \w_{i})$
            \State $\w_{i+1} \gets \w_{i} + \D_z\hat{\x}_{0|t_{i}} - \z_{i+1}$
            \State Sample $\bepsilon_i \sim \mathcal{N}(\mathbf{0}, \I_d)$
            \State $\x_{t_{i+1}} \gets \alpha_{t_{i+1}}\hat{\x}_{0|t_{i}} + \sqrt{\sigma^2_{t_{i+1}}-\eta^2\beta^2_{t_{i+1}}}\hat\bepsilon_i + \eta\beta_{t_{i+1}}\bepsilon_i$ \Comment{DDIM sampling to $t_{i+1}$}
        \EndFor
        \State $\mathbf{C} \gets \mathbf{\infty}_{(N+1) \times (N+1)}$
        \For {$i \gets 0$ to $N-1$}
            \For {$j \gets i+1$ to $N$}
                \State $\overline\x_{t_{j}|t_{i}} \gets \alpha_{t_{j}}\hat{\x}_{0|t_{i}} + \sqrt{\sigma^2_{t_{j}}-\eta^2\beta^2_{t_{j}}}\hat\bepsilon_i + \eta\beta_{t_{j}}\bepsilon_{j-1}$ 
                \Statex \Comment{DDIM sampling to $t_j$} with reusing noise $\bepsilon_{j-1}$
                \State $C_{i,j} \gets \|\overline{\x}_{t_j|t_{i}} - \x_{t_j}\|_2  + \kappa \left(\left(t_i - t_j\right) - \frac{T}{L}\right)^2$
            \EndFor
        \EndFor
        \State $\{\tau_j\}_{j=0}^{L} \gets \text{DP}(\mathbf{C}, \{t_i\}_{i=0}^{N}, N, L)$ 
        \Statex \Comment{Execute Dynamic Programming Algorithm to solve Eq.~\eqref{eq:optimModel}}
        \State \Return $\{\tau_j\}_{j=0}^{L}$
    \end{algorithmic}
\end{algorithm}

\begin{algorithm}[!t]
    \caption{Dynamic Programming Solver (DP)}
    \label{alg:dp}
    \begin{algorithmic}[1]
        \Require Cost matrix $\mathbf{C}$, Source time steps $\{t_i\}_{i=1}^N$, Source number of steps $N$, Target number of steps $L$
        \State $\mathbf{State} \gets \mathbf{\infty}_{L \times (N+1)}$, $\mathbf{Path} \gets \mathbf{0}_{L \times (N+1)}$
        \For {$j = 1$ to $N-1$}
            \State $\mathbf{State}[0, j] \gets \mathbf{C}[0, j]$
            \State $\mathbf{Path}[0, j] \gets 0$
        \EndFor
        \For {$i \gets 1$ to $L-1$}
            \For {$j \gets i+1$ to $N$}
                \For {$k \gets i-1$ to $j-1$}
                    \If{$\mathbf{State}[i-1, k] + \mathbf{C}[k, j] < \mathbf{State}[i, j]$}
                        \State $\mathbf{State}[i, j] \gets \mathbf{State}[i-1, k] + \mathbf{C}[k, j]$
                        \State $\mathbf{Path}[i, j] \gets k$
                    \EndIf
                \EndFor
            \EndFor
        \EndFor
        \State $\text{Timesteps} \gets [t_{N}]$
        \State $i \gets N$
        \For {$j \gets L-1$ down to $0$}
            \State $i \gets \mathbf{Path}[j, i]$
            \State Append $t_i$ to $\text{Timesteps}$
        \EndFor
        \State \Return $\text{reverse}(\text{Timesteps})$
    \end{algorithmic}
\end{algorithm}

\section{Experimental Results}
\label{sec:experiments}

\subsection{Experimental Setups}

\subsubsection{3D CT Reconstruction}
We evaluate the proposed framework on sparse-view CT (SV-CT) and limited-angle CT (LA-CT) tasks. The forward operator $\A$ employs the discrete Radon transform. For SV-CT, we uniformly sample 2, 4, and 8 projection angles. For LA-CT, we restrict 120 projections to a $90^\circ$ wedge. Experiments are conducted on the AAPM dataset \cite{aapm2016dataset} following standard protocols~\cite{chungdecomposed,chung2023solving}. We train the 2D diffusion prior on the training split, and designate one specific volume to compute the oracle trajectory ($N=200$) and extract the optimized schedule. Subsequent evaluations are conducted on a distinct volume, which serves as the standard single-volume benchmark commonly adopted by prior works~\cite{chungdecomposed,chung2023solving}. Although community conventions restrict the evaluation to this single 3D volume, the reconstruction actually entails hundreds of independent 2D inferences across its high-resolution slices. Therefore, reporting metrics averaged over the axial, coronal, and sagittal planes provides a reliable quantitative assessment of our method.

\subsubsection{Implementation Details}
We integrate our proposed timestep scheduling algorithm into DDS~\cite{chungdecomposed}. Since existing ODE-based optimized schedules are difficult to apply directly to stochastic conditional inverse solvers, we evaluate our approach against several widely adopted fixed schedules within the identical pipeline to ensure a fair comparison. These baselines include the uniform time schedule Uniform-$t$~\cite{song2021ddim}, the uniform half of log-SNR schedule Uniform-$\lambda$~\cite{lu2022dpmsolver}, the Quadratic schedule~\cite{song2021ddim}, the EDM schedule~\cite{karras2022elucidating}, and the Cosine schedule~\cite{nichol2021improved}. Detailed formulations for these schedules are provided in the supplementary material. To isolate the impact of the scheduling strategies, all methods utilize the identical pre-trained diffusion model, and the algorithmic hyperparameters are kept consistent with the original DDS implementation. Additionally, we compare our method against established diffusion solvers for 3D inverse problems, such as DiffMBIR~\cite{chung2023solving}, Score-Med~\cite{song2023solving}, and MCG~\cite{chung2022improving}, which typically demand thousands of NFEs to achieve competitive reconstruction quality. Following approaches~\cite{chung2023solving,chungdecomposed,chung2022improving,song2023solving}, volumetric reconstruction quality is quantitatively assessed using the Peak Signal-to-Noise Ratio (PSNR) and the Structural Similarity Index Measure (SSIM), evaluated on Axial, Coronal, and Sagittal planes. 

\subsection{Quantitative Results}
\begin{figure}
    \centering
    \includegraphics[width=1.\linewidth]{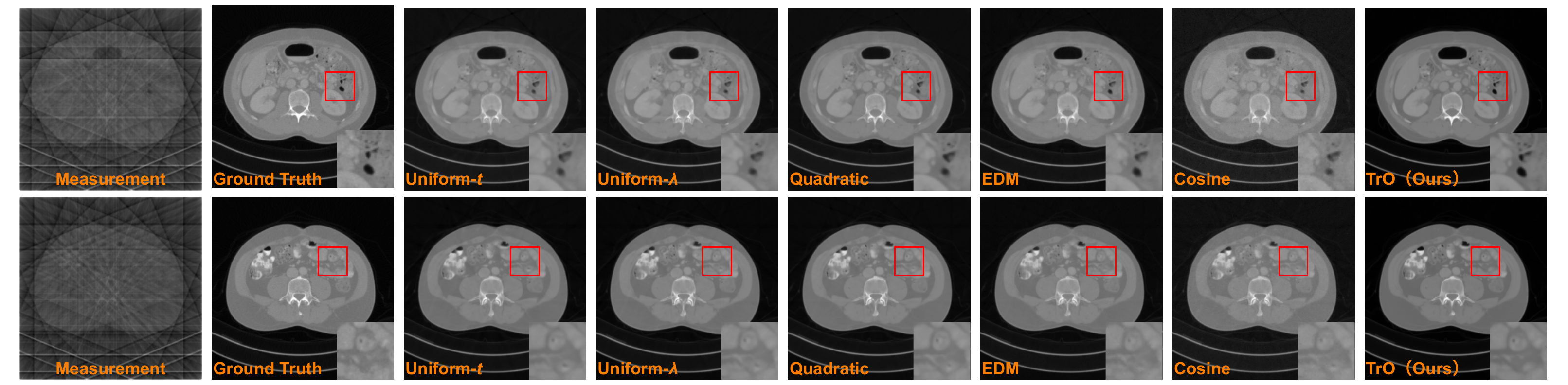}
    \caption{Qualitative comparison of 3D CT reconstruction results using different timestep schedules. Red boxes highlight the detail recovery of our method.}
\label{fig:visual_comparison}
\end{figure}

We comprehensively evaluate the volumetric reconstruction performance across SV-CT (e.g., 8-view and 4-view) and LA-CT tasks under strictly constrained computational budgets corresponding to 15, 10 and 8 NFEs. The quantitative comparisons, averaged across the axial, coronal, and sagittal planes, are summarized in Table~\ref{tab:full_ablation}. The evaluation indicates that TrO yields competitive reconstruction fidelity across the tested configurations. Notably, the advantage of our optimized schedule over predefined counterparts becomes more evident as the NFE budget decreases. While this performance gap may tend to narrow as the number of NFEs increases, these results highlight the utility of TrO in further enhancing the efficiency of the DDS framework with limited computational resources. Additionally, visual comparisons of the reconstructed tomographic slices are provided in Fig.~\ref{fig:visual_comparison}, which further corroborate the enhanced artifact suppression and detail preservation capabilities of our adaptive schedule.
\begin{table*}[tp]
\centering
\caption{Quantitative comparison of timestep schedules on AAPM across different NFEs and SV/LA-CT. Best results are highlighted in \textbf{bold} and the second-best results are \underline{underlined}.}
\label{tab:full_ablation}

\scriptsize
\renewcommand{\arraystretch}{0.80}
\setlength{\tabcolsep}{1.8pt}

\begin{adjustbox}{max width=\textwidth, max totalheight=0.88\textheight, keepaspectratio}
\begin{tabular}{cc cccccc cccccc}
\toprule
\multirow{4}{*}{Method (NFE)} & \multirow{4}{*}{Schedule} & \multicolumn{6}{c}{\textbf{8-view}} & \multicolumn{6}{c}{\textbf{4-view}} \\
\cmidrule(lr){3-8} \cmidrule(lr){9-14}
& & \multicolumn{2}{c}{Axial} & \multicolumn{2}{c}{Coronal} & \multicolumn{2}{c}{Sagittal} & \multicolumn{2}{c}{Axial} & \multicolumn{2}{c}{Coronal} & \multicolumn{2}{c}{Sagittal} \\
\cmidrule(lr){3-4} \cmidrule(lr){5-6} \cmidrule(lr){7-8} \cmidrule(lr){9-10} \cmidrule(lr){11-12} \cmidrule(lr){13-14}
& & PSNR$\uparrow$ & SSIM$\uparrow$ & PSNR$\uparrow$ & SSIM$\uparrow$ & PSNR$\uparrow$ & SSIM$\uparrow$ & PSNR$\uparrow$ & SSIM$\uparrow$ & PSNR$\uparrow$ & SSIM$\uparrow$ & PSNR$\uparrow$ & SSIM$\uparrow$ \\
\midrule

\multirow{6}{*}{DDS (8)}
& Uniform-$t$       & 37.08 & 0.940 & 37.89 & \underline{0.935}& 36.74 & \underline{0.936}& 32.09 & 0.894 & 33.05 & 0.887 & 31.13 & 0.889 \\
& Uniform-$\lambda$ & 36.68 & 0.937 & 38.03 & 0.931 & 36.94 & 0.932 & 31.99 & 0.894 & 33.14 & 0.889 & 31.16 & \underline{0.891}\\
& Quadratic         & \underline{37.58}& \underline{0.941}& \underline{38.59}& 0.931 & \underline{37.87}& 0.934 & \underline{32.66}& \underline{0.901}& \underline{33.60}& \underline{0.890}& \underline{31.57}& \underline{0.891}\\
& EDM               & 35.59 & 0.928 & 36.95 & 0.922 & 36.09 & 0.923 & 30.42 & 0.830 & 30.27 & 0.815 & 29.79 & 0.821 \\
& Cosine            & 36.60 & 0.933 & 37.64 & 0.926 & 36.70 & 0.929 & 31.77 & 0.890 & 33.02 & 0.885 & 31.24 & 0.887\\
& TrO (Ours)    & \textbf{38.53}& \textbf{0.952}& \textbf{39.78}& \textbf{0.964}& \textbf{38.62}& \textbf{0.948} & \textbf{33.24}& \textbf{0.917}& \textbf{34.66}& \textbf{0.910}& \textbf{32.78}& \textbf{0.914}\\
\midrule

\multirow{6}{*}{DDS (10)}
& Uniform-$t$       & 37.66 & 0.945 & 38.52 & 0.939 & 37.34 & 0.941 & 32.78 & 0.900 & 33.56 & 0.895 & 31.53 & 0.896 \\
& Uniform-$\lambda$ & 38.08 & 0.948 & 39.03 & 0.943 & 37.96 & 0.944 & 32.34 & 0.899 & 33.55 & 0.894 & 31.51 & 0.895 \\
& Quadratic         & \underline{38.72} & \underline{0.951} & \underline{39.67} & \underline{0.946} & \underline{38.66} & \underline{0.948} & 33.29 & 0.908 & 34.25 & 0.899 & 32.15 & 0.901\\
& EDM           & 38.38 & 0.946 & 39.37 & 0.940 & 38.61 & 0.942 & \underline{33.66} & \underline{0.921}& \underline{34.63}& \underline{0.916}& \underline{33.21}& \underline{0.917}\\
& Cosine           & 36.29 & 0.907 & 36.42 & 0.895 & 35.96 & 0.900 & 30.88 & 0.800 & 30.96 & 0.780 & 30.12 & 0.791 \\
& TrO (Ours)    & \textbf{39.33} & \textbf{0.955} & \textbf{40.36} & \textbf{0.950} & \textbf{39.46} & \textbf{0.951} & \textbf{34.12} & \textbf{0.927}& \textbf{35.37} & \textbf{0.922} & \textbf{33.86} & \textbf{0.920} \\
\midrule

\multirow{6}{*}{DDS (15)}
& Uniform-$t$       & 38.54 & 0.950 & 39.47 & 0.945 & 38.36 & 0.947 & 33.82 & 0.911 & 34.67 & 0.906 & 32.68 & 0.907 \\
& Uniform-$\lambda$ & 39.01 & 0.953 & 40.04 & 0.948 & 39.01 & 0.949 & 33.54 & 0.907 & 34.48 & 0.904 & 32.36 & 0.905 \\
& Quadratic         & 39.63 & 0.955 & 40.61 & \underline{0.951}& 39.69 & 0.952 & \underline{34.84}& 0.923 & 35.68 & 0.917 & 33.66 & 0.918 \\
& EDM           & \underline{39.82}& \underline{0.956}& \underline{40.90}& \underline{0.951}& \underline{40.06}& \underline{0.953}& \underline{34.84}& \underline{0.929}& \textbf{36.50}& \textbf{0.924}& \underline{34.49}& \underline{0.925}\\
& Cosine           & 36.63 & 0.905 & 36.69 & 0.894 & 36.31 & 0.900 & 30.68 & 0.763 & 30.58 & 0.744 & 30.03 & 0.758 \\
& TrO (Ours) & \textbf{40.02} & \textbf{0.957} & \textbf{40.96} & \textbf{0.952} & \textbf{40.14} & \textbf{0.954} & \textbf{34.99} & \textbf{0.932} & \underline{36.48}& \underline{0.923}& \textbf{34.61} & \textbf{0.927} \\
\midrule

DiffMBIR(4000)&    /    & 33.49 & 0.942 & 35.18 & 0.967 & 32.18 & 0.910 & 30.52 & 0.914 & 30.09 & 0.938 & 27.89 & 0.871 \\
Score-Med(4000)&    /     & 29.10 & 0.882 & 27.93 & 0.875 & 24.23 & 0.759 & 28.20 & 0.867 & 27.48 & 0.889 & 25.08 & 0.783   \\
MCG(4000)&     /      & 28.61 & 0.873 & 28.05 & 0.884 & 24.45 & 0.765 & 27.33 & 0.855 & 26.52 & 0.863 & 23.04 & 0.745 \\

\midrule
\midrule
\multirow{4}{*}{Method(NFE)} & \multirow{4}{*}{Schedule} & \multicolumn{6}{c}{\textbf{2-view}} & \multicolumn{6}{c}{\textbf{$90^\circ$-Angle}} \\
\cmidrule(lr){3-8} \cmidrule(lr){9-14}
& & \multicolumn{2}{c}{Axial} & \multicolumn{2}{c}{Coronal} & \multicolumn{2}{c}{Sagittal} & \multicolumn{2}{c}{Axial} & \multicolumn{2}{c}{Coronal} & \multicolumn{2}{c}{Sagittal} \\
\cmidrule(lr){3-4} \cmidrule(lr){5-6} \cmidrule(lr){7-8} \cmidrule(lr){9-10} \cmidrule(lr){11-12} \cmidrule(lr){13-14}
& & PSNR$\uparrow$ & SSIM$\uparrow$ & PSNR$\uparrow$ & SSIM$\uparrow$ & PSNR$\uparrow$ & SSIM$\uparrow$ & PSNR$\uparrow$ & SSIM$\uparrow$ & PSNR$\uparrow$ & SSIM$\uparrow$ & PSNR$\uparrow$ & SSIM$\uparrow$ \\
\midrule

\multirow{6}{*}{DDS (8)}
& Uniform-$t$       & 25.04 & 0.742 & 25.98 & 0.723 & 24.12 & 0.710 & 38.07 & 0.944 & 39.33 & 0.936 & 38.27 & 0.939 \\
& Uniform-$\lambda$ & 24.55 & 0.730 & 25.76 & 0.714 & 23.97 & 0.707 & 38.29 & \underline{0.949}& \underline{39.84}& \underline{0.942}& 38.52 & \underline{0.945}\\
& Quadratic         & \underline{25.18}& \underline{0.752}& \underline{26.02}& \underline{0.753}& \underline{24.16}& \underline{0.740}& 38.28 & 0.944 & 39.73 & \underline{0.942}& \underline{38.60} & \textbf{0.946}\\
& EDM               & 21.50 & 0.606 & 22.03 & 0.634 & 21.30 & 0.615 & 36.71 & 0.941 & 38.26& 0.934 & 36.64 & 0.937 \\
& Cosine            & 23.39 & 0.705 & 25.52 & 0.752 & 23.11 & 0.704 & \underline{38.36}& 0.947 & 39.62 & 0.940 & {38.57}& \textbf{0.946}\\
& TrO (Ours)    & \textbf{26.23}& \textbf{0.798}& \textbf{26.74}& \textbf{0.789}& \textbf{24.74}& \textbf{0.803}& \textbf{38.82}& \textbf{0.952}& \textbf{40.32}& \textbf{0.945}& \textbf{39.21}& \textbf{0.946}\\
\midrule

\multirow{6}{*}{DDS (10)}
& Uniform-$t$       & 25.52 & 0.731 & 26.30 & 0.712 & 24.55 & 0.720 & 38.52 & 0.948 & 39.85 & 0.941 & 38.77 & 0.943 \\
& Uniform-$\lambda$ & 25.25 & 0.741 & 25.85 & 0.724 & 24.26 & 0.729 & 38.61 & \underline{0.950}& 40.14 & 0.944 & 38.91 & \underline{0.946}\\
& Quadratic         & \underline{26.05}& \underline{0.781}& \underline{27.03}& \underline{0.798}& \underline{24.73}& \underline{0.796}& 38.54 & 0.947 & \underline{40.17}& \underline{0.945}& 38.87 & 0.945 \\
& EDM           & 23.27 & 0.728 & 23.68 & 0.762 & 22.50 & 0.739 & 37.76 & 0.946 & 39.54 & 0.940 & 38.06 & 0.943\\
& Cosine           & 24.06 & 0.558 & 24.34 & 0.525 & 23.45 & 0.546 & \underline{38.68}& \underline{0.950}& 40.07 & 0.943& \underline{38.93}& \underline{0.946}\\
& TrO (Ours)    & \textbf{26.70}& \textbf{0.811}& \textbf{27.60}& \textbf{0.819}& \textbf{25.36}& \textbf{0.818}& \textbf{39.17}& \textbf{0.954}& \textbf{40.51}& \textbf{0.947}& \textbf{39.47}& \textbf{0.949}\\
\midrule

\multirow{6}{*}{DDS (15)}
& Uniform-$t$       & 25.93 & 0.681 & 26.32 & 0.667 & 24.77 & 0.680 & \underline{39.00}& \underline{0.951}& \underline{40.35}& \underline{0.945}& \underline{39.28}& \underline{0.947}\\
& Uniform-$\lambda$ & 25.52 & 0.702 & 26.01 & 0.687 & 24.48 & 0.697 & 38.80 & 0.950 & 40.30 & 0.944 & 39.14 & 0.946 \\
& Quadratic         & \underline{26.93}& \underline{0.813}& \underline{28.01}& \underline{0.821}& \underline{25.27}& \underline{0.827}& 38.88 & \textbf{0.952}& 40.33 & 0.944 & 39.22 & \underline{0.947}\\
& EDM           & 26.37 & 0.802 & 26.41 & 0.809 & 24.61 & 0.803 & 37.93 & 0.944 & 39.69 & 0.937 & 38.33 & 0.940 \\
& Cosine           & 24.04 & 0.491 & 24.03 & 0.463 & 23.34 & 0.489 & 38.97 & 0.950 & 40.32 & 0.944 & 39.25 & 0.946 \\
& TrO (Ours)    & \textbf{27.29}& \textbf{0.820}& \textbf{28.24}& \textbf{0.833}& \textbf{25.74}& \textbf{0.832}& \textbf{39.13}& \textbf{0.952}& \textbf{40.52}& \textbf{0.946}& \textbf{39.46}& \textbf{0.948}\\
\midrule

DiffMBIR(4000)   &  / & 24.11 & 0.810 & 23.15 & 0.841 & 21.72 & 0.766 & 34.92 & 0.956 & 32.48 & 0.947 & 28.82 & 0.832 \\
Score-Med(4000) &   /   & 24.07 & 0.808 & 23.70 & 0.822 & 20.95 & 0.720 & 27.80 & 0.852 & 25.69 & 0.869 & 22.03 & 0.735 \\
MCG(4000)        &  /  & 24.69 & 0.821 & 23.52 & 0.806 & 20.71 & 0.685 & 26.01 & 0.838 & 24.55 & 0.823 & 21.59 & 0.706 \\

\bottomrule
\end{tabular}
\end{adjustbox}
\end{table*}

\subsection{Further Analysis}
\label{sec:FurtherAnalysis}
\begin{figure}
    \centering
    \includegraphics[width=0.85\linewidth]{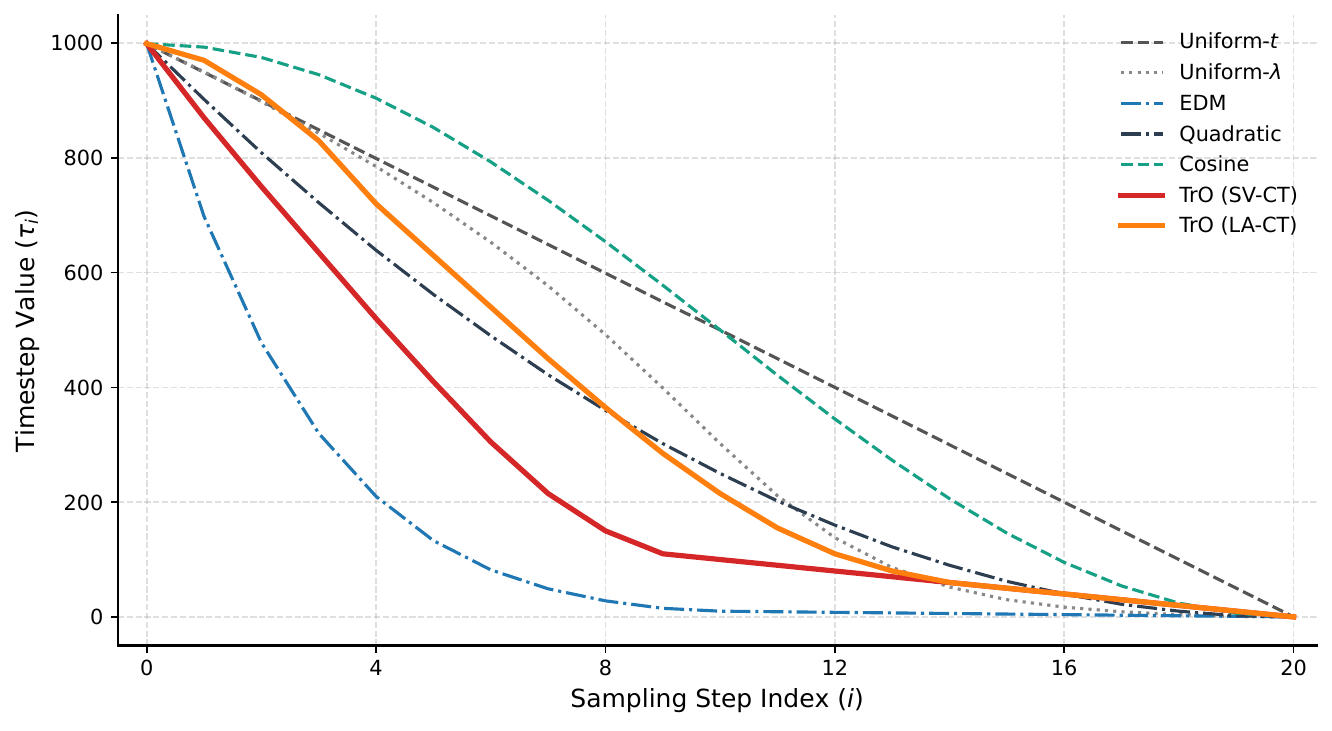}
    \caption{Visualization of timestep allocations across different scheduling strategies. The graph illustrates the temporal evolution of continuous time $\tau_i$ over discrete evaluation steps $i$.}
\label{fig:timestep_evolution}
\end{figure}

Fig.~\ref{fig:timestep_evolution} illustrates the actual temporal evolution of various scheduling strategies. The quantitative results in Table~\ref{tab:full_ablation} indicate that the relative performance of static timestep schedules is highly dependent on the specific measurement operator and the available NFEs. For instance, while the Cosine schedule exhibits a noticeable performance drop in SV-CT tasks, it demonstrates relatively better stability in LA-CT settings compared to the EDM and Quadratic schedules. 
\subsubsection{Ablation Study}
We conduct an ablation study on the AAPM dataset to investigate the impact of the regularization hyperparameter $\kappa$ in Eq.~\eqref{eq:cost_matrix} for 8-view SV-CT reconstruction task. The reported PSNR and SSIM are averaged across the axial, coronal, and sagittal planes. As shown in Table~\ref{tab:ablation_kappa}, the optimal $\kappa$ varies depending on the NFEs. Generally, a larger NFE requires a correspondingly larger $\kappa$ to enforce a similar level of constraint, which stems from the differences in $N/L$. Furthermore, the empirical results indicate that the introduction of the regularization term effectively prevents the generated schedule from degenerating into extreme timestep allocations, which enhances the overall reconstruction performance.
\begin{table}[tp]
\caption{Ablation study of $\kappa$ for the 8-view SV-CT task on AAPM. Best results are highlighted in \textbf{bold} and the second-best results are \underline{underlined}.}
\label{tab:ablation_kappa}
\centering
\begin{tabular}{ccccccc}
\hline
NFE                & Metric & $\kappa=0$ & $\kappa=1$ & $\kappa=5$ & $\kappa=10$ & Uniform-$t$\\ \hline
\multirow{2}{*}{8} &  PSNR$\uparrow$ & 38.42 & \textbf{38.98} & \underline{38.65} & 37.64 & 37.24 \\
                   &  SSIM$\uparrow$ & 0.951 & \textbf{0.955} & \underline{0.953} & 0.946 & 0.942\\ \hline
\multirow{2}{*}{10} &  PSNR$\uparrow$ & 39.32 & 39.64 & \textbf{39.72} & \underline{38.96} & 37.84 \\
                   &  SSIM$\uparrow$ & 0.943 & \underline{0.949} & \textbf{0.952} & 0.948 & 0.942\\ \hline
\multirow{2}{*}{15} &  PSNR$\uparrow$ & 39.63 & 39.77 & \textbf{40.37} & \underline{39.83} & 38.80\\
                   &  SSIM$\uparrow$ & 0.951 & 0.951 & \textbf{0.954} & \underline{0.952} & 0.947 \\ \hline
\end{tabular}
\end{table}

\subsubsection{Timestep Analysis}
We hypothesize that this divergence originates from the frequency reconstruction dynamics of diffusion models. The early stages of the reverse diffusion process primarily synthesize low-frequency global structures, while the later stages refine high-frequency details. In SV-CT, the uniformly distributed projections provide reliable global low-frequency constraints. Consequently, computational resources are better allocated to the later stages (e.g., as in EDM and Quadratic) for high-frequency artifact suppression. Conversely, LA-CT concentrates projections within a narrow region, introducing severe structural artifacts and unreliable low-frequency information in the unmeasured wedges. Allocating insufficient steps to the early phase implicitly over-trusts these distorted low-frequency measurements, severely impeding late-stage optimization. The Cosine schedule mitigates this by densely populating the early stages, providing the generative prior with sufficient capacity to synthesize the missing reliable low-frequency structures.

These observations reveal that a single fixed timestep allocation is sub-optimal across diverse inverse problems. Notably, our proposed TrO framework effectively captures this physical discrepancy. As clearly depicted in Fig.~\ref{fig:timestep_evolution}, TrO adaptively shifts the step allocation toward the earlier stages for the LA-CT task compared to the SV-CT task. By continuously evaluating the actual SDE truncation error, TrO dynamically extracts task-specific and budget-aware temporal configurations, thereby ensuring consistent reconstruction fidelity across varying degradation scenarios.

\section{Conclusion}
\label{sec:conclusion}
In this paper, we present Tracing the Oracle (TrO), a plug-and-play framework that formulates timestep scheduling for diffusion-based 3D inverse problems as a discrete shortest-path optimization problem. To quantify the trajectory discretization error, we propose a noise reuse mechanism that isolates the deterministic truncation error within the SDE sampling procedure. By leveraging dynamic programming, TrO extracts an optimized timestep schedule from a densely sampled oracle trajectory. Evaluations on sparse-view and limited-angle CT reconstruction demonstrate that our proposed approach improves reconstruction fidelity, particularly under limited computational budgets. By reducing the performance gap between few-step sampling and high-NFE solvers without additional inference overhead, TrO facilitates the efficient and robust deployment of diffusion models in 3D medical imaging applications.

\appendices
\section{Timestep Schedule}
To evaluate the performance of our proposed Tracing the Oracle (TrO) framework, we compare it against several widely adopted fixed timestep schedules. All schedules partition the continuous time interval $[\epsilon, T]$ into $N$ discrete steps $\{t_n\}_{n=0}^N$, where $t_0 = T$ and $t_N = \epsilon\approx0$. The formulations for these schedules are summarized as follows:

\subsection{Uniform-\texorpdfstring{$t$}{t} Schedule}
The Uniform-$t$ schedule splits the time interval $[\epsilon, T]$ into $N$ sub-intervals of equal length \cite{song2021ddim}. The $n$-th timestep is defined as:
\begin{equation}
    t_n = T + \frac{n}{N}(\epsilon - T), \quad n = 0, 1, \dots, N.
\end{equation}

\subsection{Uniform-\texorpdfstring{$\lambda$}{lambda} Schedule}
The Uniform-$\lambda$ schedule is based on the half of log-signal-to-noise ratio (SNR), defined as $\lambda_t = \log(\alpha_t / \sigma_t)$. This scheme uniformly discretizes the interval $[\lambda_T, \lambda_\epsilon]$ and maps the values back to the time domain \cite{lu2022dpmsolver}:
\begin{equation}
    t_n = t_\lambda \left( \lambda_T + \frac{n}{N}(\lambda_\epsilon - \lambda_T) \right),
\end{equation}
where $t_\lambda(\cdot)$ denotes the inverse function of $\lambda_t$.

\subsection{EDM Schedule}
The EDM schedule employs a non-linear discretization strategy based on the variable substitution $\kappa_t = \sigma_t / \alpha_t$ \cite{karras2022elucidating}. It uniformly discretizes $\kappa_t^{1/\rho}$ for a positive integer $\rho$:
\begin{equation}
    t_n = t_\kappa \left( \left( \kappa_T^{1/\rho} + \frac{n}{N}(\kappa_\epsilon^{1/\rho} - \kappa_T^{1/\rho}) \right)^\rho \right),
\end{equation}
where $t_\kappa(\cdot)$ is the inverse function of $\kappa_t$. Following the standard implementation, we set $\rho = 7$ in our experiments.

\subsection{Quadratic Schedule}
The Quadratic schedule allocates more steps toward the smaller noise levels (near $t=0$) to refine high-frequency details \cite{song2021ddim}. The timesteps are determined by:
\begin{equation}
    t_n = \left( \sqrt{T} + \frac{n}{N}(\sqrt{\epsilon} - \sqrt{T}) \right)^2.
\end{equation}

\subsection{Cosine Schedule}
The Cosine schedule is designed to ensure a more gradual loss of information during the forward diffusion process \cite{nichol2021improved}. It is typically defined through the cumulative noise schedule $\bar{\alpha}_t$:
\begin{equation}
    \bar{\alpha}_t = \frac{f(t)}{f(0)}, \quad f(t) = \cos \left( \frac{t/T + s}{1 + s} \cdot \frac{\pi}{2} \right)^2,
\end{equation}
where $s$ is a small offset (e.g., $s=0.008$) to prevent singularities at $t=0$.
\section{Detailed DDS}
\subsection{ADMM}

To address the highly ill-posed 3D CT inverse problem, we incorporate total variation (TV) regularization along the z-axis to leverage cross-slice structural priors:
\begin{equation}
    \min_{\x} \frac{1}{2} \|\y - \A\x\|_2^2 + \zeta \|\D_z \x\|_1,
\end{equation}
where $\A$ is the Radon transform operator, $\y$ denotes the projection measurements, $\D_z$ represents the finite difference operator along the axial direction, and $\zeta$ is the regularization parameter.

\subsubsection{Constraint Reformulation}
By introducing an auxiliary variable $\z \in \mathbb{R}^d$, we convert the unconstrained $L_1$ problem into an equivalent constrained optimization framework:
\begin{equation}
    \min_{\x, \z} \frac{1}{2} \|\y - \A\x\|_2^2 + \zeta \|\z\|_1, \quad \text{s.t. } \z = \D_z \x.
\end{equation}
The corresponding Augmented Lagrangian function is defined as:
\begin{equation}
    L_\rho(\x, \z, \w) = \frac{1}{2} \|\y - \A\x\|_2^2 + \zeta \|\z\|_1 + \frac{\rho}{2} \|\D_z \x - \z + \w\|_2^2,
\end{equation}
where $\w$ is the scaled dual variable and $\rho > 0$ denotes the ADMM penalty parameter.

\subsubsection{Iterative Updates}
The ADMM algorithm proceeds by alternating the minimization of $L_\rho$ with respect to each variable:

\paragraph{Primal Variable $\x$-update:}
Fixing $\z_j$ and $\w_j$, the update for $\x$ is obtained by solving:
\begin{equation}
    \x_{j+1} = \arg\min_{\x} \frac{1}{2} \|\y - \A\x\|_2^2 + \frac{\rho}{2} \|\D_z \x - \z_j + \w_j\|_2^2.
\end{equation}
Taking the gradient with respect to $\x$ and setting it to zero yields the following linear system:
\begin{equation}
    (\A^\top \A + \rho \D_z^\top \D_z) \x_{j+1} = \A^\top \y + \rho \D_z^\top (\z_j - \w_j).
\end{equation}
In our framework, this high-dimensional system is solved iteratively using the Conjugate Gradient (CG) method.

\paragraph{Auxiliary Variable $\z$-update:}
Fixing $\x_{j+1}$ and $\w_j$, the update for $\z$ is given by:
\begin{equation}
    \z_{j+1} = \arg\min_{\z} \zeta \|\z\|_1 + \frac{\rho}{2} \|\D_z \x_{j+1} - \z + \w_j\|_2^2.
\end{equation}
This proximal subproblem has a closed-form solution via the soft-thresholding operator $\mathcal{S}_{\tau}(\cdot)$:
\begin{equation}
    \z_{j+1} = \mathcal{S}_{\zeta/\rho} (\D_z \x_{j+1} + \w_j),
\end{equation}
where $\mathcal{S}_{\tau}(u) = \text{sign}(u) \cdot \max(|u| - \tau, 0)$.

\paragraph{Dual Variable $\w$-update:}
The dual variable is updated to enforce the linear constraint:
\begin{equation}
    \w_{j+1} = \w_j + \D_z \x_{j+1} - \z_{j+1}.
\end{equation}

To ensure computational efficiency during the reverse diffusion process, we perform only a single ADMM iteration per sampling step and share the variables $\{\z, \w\}$ across timesteps, a technique known as fast variable sharing.
\subsection{CG}
The detailed implementation for the conjugate gradient (CG) is given in Algorithm~\ref{alg:cg_full}.
\begin{algorithm}[H]
\caption{Conjugate Gradient (CG)}
\label{alg:cg_full}
\begin{algorithmic}[1]
\Require Symmetric positive definite matrix $\A$, right-hand side $\mathbf{b}$, initial estimate $\x_0$, number of iterations $M$
\State $\boldsymbol{r}_0 \gets \mathbf{b} - \A\x_0$
\State $\mathbf{p}_0 \gets \boldsymbol{r}_0$
\For{$k = 0$ to $M-1$}
    \State $\alpha_k \gets \frac{\boldsymbol{r}_k^\top \boldsymbol{r}_k}{\mathbf{p}_k^\top \A \mathbf{p}_k}$
    \State $\x_{k+1} \gets \x_k + \alpha_k \mathbf{p}_k$
    \State $\boldsymbol{r}_{k+1} \gets \boldsymbol{r}_k - \alpha_k \A \mathbf{p}_k$
    \State $\beta_k \gets \frac{\boldsymbol{r}_{k+1}^\top \boldsymbol{r}_{k+1}}{\boldsymbol{r}_k^\top \boldsymbol{r}_k}$
    \State $\mathbf{p}_{k+1} \gets \boldsymbol{r}_{k+1} + \beta_k \mathbf{p}_k$
\EndFor
\State \Return $\x_M$
\end{algorithmic}
\end{algorithm}

\subsection{Full DDS in 3D CT Reconstruction}
After obtaining the optimized timesteps $\{\tau_i\}_{i=0}^{L}$, we utilize this schedule with DDS~\cite{chungdecomposed} to achieve 3D reconstruction. This step is analogous to computing the oracle trajectory in the TrO algorithm, with the complete procedure summarized in Algorithm~\ref{alg:full_dds}.
\begin{algorithm}[H]
    \caption{DDS}
    \label{alg:full_dds}
    \begin{algorithmic}[1]
        \Require Target timesteps $\{\tau_i\}_{i=0}^{L}$, $\{\alpha_{\tau_i}\}_{i=0}^{L}$, $\{\sigma_{\tau_i}\}_{i=0}^{L}$ and $\{\beta_{\tau_i}\}_{i=0}^{L}$,  $\A$, $\y$, $\D_z$, $\rho$, $\mathbf{S}_{\zeta/\rho}$, $M$, $\eta$
        \State Sample $\x_{\tau_{0}} \sim \mathcal{N}(\mathbf{0}, \mathbf{I}_d)$
        \State $ \z_0 \gets \mathbf{0}_{d}, \w_0 \gets \mathbf{0}_d$
        \For {$i \gets 0$ to $L-1$}
            \State $\hat\bepsilon_i \gets \bepsilon_\theta(\x_{\tau_i}, \tau_i)$
            \State $\x_{0|\tau_{i}} \gets (\x_{\tau_i} - \sigma_{\tau_i}\hat\bepsilon_i)/\alpha_{\tau_i}$ \Comment{Tweedie denoising}
            \State $\A_\text{CG} \gets \A^\top\A + \rho\D_z^\top\D_z$
            \State $\mathbf{b}_\text{CG} \gets \A^\top \y + \rho\D_z^\top(\z_{i}-\w_{i})$
            \State $\hat{\x}_{0|\tau_{i}} \gets \text{CG}(\A_\text{CG}, \mathbf{b}_\text{CG}, \x_{0|\tau_{i}}, M)$ \Comment{Data consistency}
            \State $\z_{i+1} \gets \mathbf{S}_{\zeta/\rho}(\D_z\hat{\x}_{0|\tau_{i}} + \w_{i})$
            \State $\w_{i+1} \gets \w_{i} + \D_z\hat{\x}_{0|\tau_{i}} - \z_{i+1}$
            \State Sample $\bepsilon_i \sim \mathcal{N}(\mathbf{0}, \I_d)$
            \State $\x_{\tau_{i+1}} \gets \alpha_{\tau_{i+1}}\hat{\x}_{0|\tau_{i}} + \sqrt{\sigma^2_{\tau_{i+1}}-\eta^2\beta^2_{\tau_{i+1}}}\hat\bepsilon_i + \eta\beta_{\tau_{i+1}}\bepsilon_i$ \Comment{DDIM sampling to $\tau_{i+1}$}
        \EndFor
        \State \Return $\x_{\tau_L}$
    \end{algorithmic}
\end{algorithm}

\section{Details of Dynamic Programming}

We provide additional details of the dynamic programming solver used in Algorithm~\ref{alg:dp}. Let $\mathbf{State}[s,j]$ denote the minimum accumulated cost of reaching timestep $t_j$ from $t_0$ using exactly $s+1$ transitions. The corresponding predecessor index is stored in $\mathbf{Path}[s,j]$. The initialization is given by
\begin{equation}
    \mathbf{State}[0,j] = C_{0,j}, \quad \mathbf{Path}[0,j] = 0,
    \quad j=1,\dots,N.
\end{equation}
For $s=1,\dots,L-1$, the recurrence is
\begin{equation}
    \mathbf{State}[s,j] =
    \min_{k<j} \left\{ \mathbf{State}[s-1,k] + C_{k,j} \right\},
\end{equation}
and $\mathbf{Path}[s,j]$ records the index $k$ that attains the minimum. After computing $\mathbf{State}[L-1,N]$, the optimized timestep indices are obtained by backtracking from $N$ through $\mathbf{Path}$. This procedure enumerates all valid predecessor nodes at each transition count, and therefore returns the global optimum of Eq.~\eqref{eq:optimModel} under the pre-computed cost matrix $\mathbf{C}$. The time complexity is $\mathcal{O}(LN^2)$ and the memory complexity is $\mathcal{O}(LN)$.

\newpage
\bibliographystyle{IEEEtran}
\bibliography{references}

\end{document}